\documentclass{article}
\usepackage[colorlinks,
            linkcolor=red,
            anchorcolor=blue,
            citecolor=blue
            ]{hyperref}
\usepackage{nanostyle}
\usepackage{nanopage}

\usepackage[round,sort]{natbib}

\ifdefined\final
\usepackage[disable]{todonotes}
\else
\usepackage[textsize=tiny]{todonotes}
\fi
\usepackage{booktabs}
\usepackage{multirow}
\usepackage{fancyhdr}
\title{\huge Robust Layerwise Scaling Rules by Proper Weight Decay Tuning}
\renewcommand{\undertitle}{}

\author
{
    Zhiyuan Fan\footnotemark[1]\footnotetext[1]{Department of Electrical Engineering and Computer Science, MIT, Cambridge, MA, USA; email: fanzy@mit.edu}
    ~~~
    Yifeng Liu\footnotemark[2]\footnotetext[2]{Department of Computer Science, UCLA, CA, USA; email: liuyifeng@g.ucla.edu} 
	~~~
    Qingyue Zhao\footnotemark[3]\footnotetext[3]{Department of Computer Science, UCLA, CA, USA; email: zhaoqy24@g.ucla.edu} 
    ~~~
    Angela Yuan\footnotemark[4]\footnotetext[4]{Department of Computer Science, UCLA, CA, USA; email: hzyuan@cs.ucla.edu}
    ~~~
	Quanquan Gu\footnotemark[5]\footnotetext[5]{Corresponding author: Department of Computer Science, UCLA, CA, USA; email: qgu@cs.ucla.edu}
}

\date{}

\begin{document}

\maketitle

\begin{abstract}
Empirical scaling laws prescribe how to allocate parameters, data, and compute, while maximal-update parameterization ($\mu$P) enables learning-rate transfer across widths by equalizing early-time update magnitudes. However, in modern scale-invariant architectures, training quickly enters an optimizer-governed steady state where normalization layers create backward scale sensitivity and the effective learning rate becomes width dependent, degrading $\mu$P transfer. We address this by introducing a weight-decay scaling rule for AdamW that preserves sublayer gain across widths. Empirically, the singular-value spectrum of each matrix parameter scales in norm as $\sqrt{\eta/\lambda}$ with an approximately invariant shape; under width scaling $d$, we observe that the top singular value scales approximately as $\sqrt{\eta/\lambda}\cdot d^{0.75}$. Combining this observation with the $\mu$P learning-rate rule $\eta_2\propto d^{-1}$ for matrix-like parameters implies an empirical weight-decay scaling rule $\lambda_2\propto \sqrt{d}$ that approximately keeps sublayer gains width invariant. Together with vector-like parameters trained at $\eta_1=\Theta_d(1)$ and $\lambda_1=0$, this yields \emph{zero-shot} transfer of both learning rate and weight decay from proxy to target widths, removing per-width sweeps. We validate the rule on LLaMA-style Transformers and in a minimal synthetic setting, and we provide a simple diagnostic, matching top singular values, to check sublayer-gain invariance. Our results extend $\mu$P beyond the near-init regime by explicitly controlling steady-state scales set by the optimizer, offering a practical recipe for width-robust hyperparameter transfer under AdamW.
\end{abstract}

% \todoq{it is better to add Qingyue Zhao and Huizhuo Yuan as 3rd and 4th authors, as they have contributed to this project over the summer.}

\section{Introduction}

Over the past few years, \textbf{empirical scaling laws} have emerged as a guiding principle for developing ever-larger language models. A growing body of work demonstrates that test loss often follows simple power-law relationships with respect to model size, dataset size, and compute budget. These regularities provide nearly closed-form prescriptions for distributing resources: how many parameters to allocate, how much data to train on, and even which learning-rate schedule to adopt for compute-efficient training \citep{kaplan2020scaling,hoffmann2022training}. Initially derived for GPT-style Transformers and later refined under compute-optimal training regimes, these laws now serve as a foundation for many large-scale model design practices.

\textbf{Maximal-update Parameterization} ($\mu$P) \citep{yang2022tensor} complements such global scaling insights by analyzing the dynamics of individual updates. Its central principle is that as a model widens, the rate of change of each parameter tensor should remain invariant. Specifically, under suitable initialization, vector-like parameters (embeddings, LayerNorm gains, biases) should maintain a constant learning rate, while matrix parameters in attention and feed-forward layers should have learning rates that scale inversely with width. This formulation makes the optimal learning rate largely independent of model dimension, enabling one to tune it on smaller proxy models and reuse it directly for much larger counterparts.

Nevertheless, the analysis of $\mu$P primarily captures the \emph{early-time} behavior, when parameters stay close to initialization and their magnitudes are determined largely by it. In modern scale-invariant architectures, training dynamics soon reach a steady state where weight directions continue to evolve, but norms remain roughly stable due to implicit or explicit regularization. In this regime, the governing scales are dictated more by the \textit{optimizer} than the initialization \citep{kosson2023rotational,defazio2025gradients}, extending beyond the tensor-program assumptions underlying $\mu$P.

Because these steady states depend on model width, so too do the layerwise output scales. Although normalization layers enforce approximate forward scale-invariance, they introduce backward scale-sensitivity. For homogeneous normalizers such as LayerNorm or BatchNorm (without bias terms), scaling the pre-normalized activations by a factor $\alpha$ leaves the normalized output unchanged, yet the gradient with respect to the input scales inversely as $1/\alpha$ by the chain rule \citep{santurkar2018does,xu2019understanding}. Consequently, width-dependent activation magnitudes yield width-dependent gradient magnitudes, rendering the \textit{effective learning rate} scale-dependent. As a result, even if $\mu$P achieves perfect early-time matching, transfer from proxy to target models can degrade once optimization enters this steady regime. As a result, we need to tune weight decay to make sublayer gain invariant across different model width. 

In this paper, we resolve this problem by introducing a proper weight decay scaling rule for $\mu$P. Our contributions are:
\begin{itemize}
    \item We inspect the singular value spectrum of weight matrices under the steady state of AdamW training. It is observed that the singular value spectrum of each weight matrix grows in proportion to $\sqrt{\eta/\lambda}$, while the shape of the spectrum remains nearly unchanged. The proportional scaling of the learning rate and weight decay preserves the sublayer gain.

    \item When scaling up the model width $d$, we observe that the top singular value magnitude approximately scales with $\sqrt{\eta/\lambda} \cdot d^{0.75}$. Together with the learning rate scaling $\eta_2 \propto d^{-1}$ for matrix-like parameters, maintaining sublayer gain invariance requires scaling the weight decay of matrix-like parameters as $\lambda_2 \propto \sqrt{d}$. Combined with fixing the vector-like sublayers (i.e., embedding layers and RMSNorm blocks) to a learning rate of $\eta_1 = \Theta_d(1)$ and weight decay $\lambda_1 = 0$, we show that the new hyperparameter scheme achieves both optimal learning rate and optimal weight decay transfer at the same time.

    \item Finally, we present an illustrative model using synthetic data in which the $\sqrt{d}$ weight decay scaling rule can be observed. Since the synthetic data is purely random, this suggests that the weight decay scaling rule is an inherent property of the model architecture, shedding light on potentially more fine-grained layerwise scaling rules for future work.
\end{itemize}

\section{Related Work}

\subsection{Hyperparameter Transfer}

Empirical scaling rules provide global prescriptions for allocating hyperparameters, data, and compute, and have repeated shown near power-law regularities across modalities and have repeated shown near power-law regularities across modalities and architectures \citep{hestness2017deep,kaplan2020scaling,henighan2020scaling,hoffmann2022training,bjorck2024scaling,li2025predictablescalei}. While these results guide \emph{what} to scale, they say less about \emph{how} to transport tuned hyperparameters across model widths. In practice, this gap has been bridged either by black-box search (e.g., Hyperband/ASHA, BOHB, and modern HPO frameworks) \citep{jamieson2016non,li2018hyperband,perrone2018scalable,falkner2018bohb,akiba2019optuna,horvath2021hyperparameter} or by phenomenological heuristics.

A principled alternative is the \textbf{Tensor Program} view and \textbf{Maximal-update Parameterization} ($\mu$P), which unify Standard Parameterization \citep{glorot2010understanding,he2015delving}, Neural Tangent style scalings \citep{jacot2018neural}, and Mean-Field limits \citep{chizat2018global,mei2018mean,sirignano2020mean,rotskoff2022trainability}. The core prescription of $\mu$P is that to set hyperparameters such that the update magnitudes of each tensor family should be width-invariant, yielding a learning-rate split: vector-like parameters keep constant learning rates, while matrix-like parameters scale their learning rate by inversely to the model width $d$, enabling \emph{$\mu$Transfer} of tuned hyperparameters from the proxy to a target model of larger model width \citep{yang2020feature,yang2022tensor}. Empirically, $\mu$Transfer has been observed across architectures and optimizers (SGD/Adam) and at LLM scale \citep{lingle2024large,meta2024llama4,haas2024effective,dey2024sparse}. Subsequent works extended the framework (e.g., depth-wise transfer and spectral perspectives), further clarifying when early-time dynamics align across widths \citep{yang2023spectral,yang2023tensor,TensorProgramVI}.

However, despite its success, the current Tensor Program theory and the analyses of $\mu$P primarily characterize the \emph{near-initialization} regime, where the total number of effective update steps is small comparing to the model width and the initial magnitudes dominate \citep{yang2020feature,golikov2022non,yang2023spectral,yang2023tensor,chen2025global}. In modern pretraining, however, optimization typically runs for longer timestep much larger than the model width $d$ \citep{achiam2023gpt,liu2024deepseek}. In this long-horizon regime, even for linear activations the existing theory remains insufficient to describe the terminal implicit bias or generalization \citep{bordelon2022self,chizat2024infinite,bordelon2025deep}.

Furthermore, scale-invariant architectures interact with normalization in a way that creates width-dependent \emph{effective learning rates}: non-affine normalizers (BatchNorm/LayerNorm/RMSNorm without bias) preserve forward-scale invariance but introduce backward scale \emph{sensitivity}, since gradients through the normalizer scale inversely with the pre-normalized activation magnitude \citep{santurkar2018does,xu2019understanding}. As training leaves the near-init regime, norms stabilize and optimizer dynamics set the effective scale \citep{kosson2023rotational,defazio2025gradients}, so layerwise output scales (and hence gradients) become width-dependent even if early-time $\mu$P matching is perfect. This motivates width-aware \emph{weight decay} design to preserve sublayer gain across widths, the central focus of our work.

\subsection{Weight Decay Scaling Rule}

The original analysis of maximal-update \citep{yang2023tensor} produces identical predictions for any weight-decay scaling rule with $\lambda=\cO(d)$. Recently, \citet{wang2024set} advocated a linear rule $\lambda=\Theta(d)$ for AdamW, arguing that the effective shrinkage $(1-\eta\lambda)$ should not vary with $d$. Empirical learning-rate transfer studies partially corroborate this intuition: fixed $\lambda$ deteriorates transfer as width grows, while larger $\lambda$ can recover it \citep{wang2024set,lingle2024large}. Variants of linear scaling have also appeared in low-precision or variant-$\mu$P settings when measuring LR transfer via train/val loss \citep{blake2024u,narayan2025mu}, and heuristic layerwise arguments have been used to justify linear scaling for hidden matrices \citep{dey2025don}.

Beyond scaling, there is a active line of work on weight decay and its role in generalization. Decoupled weight decay was introduced to separate $L_2$ regularization from the adaptive update \citep{Adamw}. Subsequent analyses and measurements have examined norm dynamics, effective learning rates, and rotational equilibria in scale-invariant networks trained with SignGD-like methods (a family that includes Adam/AdamW) \citep{kosson2023rotational,d2024we,xiao2024rethinking,kobayashi2024weight,zhou2024towards}. Particularly relevant to us, \citet{d2024we} observed SGD generalization optima along curves approximately satisfying $\eta \propto 1/\lambda$ in under-training regimes. \citet{kosson2023rotational} further argued that in steady state the scale-invariant parameter norms track $\sqrt{\eta/\lambda}$ and per-step directional rotation scales like $\sqrt{\eta\lambda}$, a picture consistent with optimizer-governed steady-state dynamics rather than initialization-dictated ones.

Orthogonal lines explore \emph{weight-decay schedules} over time \citep{xie2023overlooked,jacobs2025mirror} and extrapolate decay rules to non-AdamW optimizers and their analyses \citep{pethick2025training,pethick2025generalized,wen2025fantastic,li2019exponential,sun2025investigating}. In contrast, we target the specific question raised by the tension above: what width-dependent $\lambda$ makes sublayer gains invariant under AdamW \emph{and} preserves $\mu$Transfer? Our empirical and synthetic analyses indicate that, when combined with the standard $\eta_2\propto d^{-1}$ for matrix-like parameters and $\eta_1=\Theta_d(1)$ with $\lambda_1=0$ for vector-like parameters, setting the matrix-parameter decay as $\lambda_2 \propto \sqrt{d}$ keeps singular-value scales (and thus sublayer gains) width-invariant in the optimizer-determined steady state.

\section{Preliminaries}

\subsection{General Notations}
We use lowercase boldface letters such as $\vv$ to denote vectors, and uppercase boldface letters such as $\mW$ to denote matrices. For a vector $\vv \in \RR^d$, the root-mean-square (RMS) norm is defined as $\|\vv\|_{\rms} \triangleq \|\vv\|_2 / \sqrt{d}$. Similarly, the RMS norm of a matrix $\mW$ is defined as the RMS norm of its vectorization. The operator norm of a matrix $\mW$ is defined as $\|\mW\|_{\mathrm{op}} \triangleq \sup_{\vx \in \RR^d} \|\vy\|_{\rms} / \|\vx\|_{\rms}$.The notation $|\vv|$ indicates the element-wise absolute value, while $\vx \odot \vy$ and $\vx \oslash \vy$ represent element-wise multiplication and division of tensors $\vx$ and $\vy$, respectively. The notation $\vx^{\odot k}$ stands for element-wise exponentiation with exponent $k$, and $\sqrt{\vx} \triangleq \vx^{\odot 1/2}$. The shorthand $\llbracket k \rrbracket \triangleq \{1, 2, \dots, k\}$ denotes an index set, and $\emptyset$ represents the empty set. The logarithm $\log x$ is taken in base $2$, while $\ln x$ refers to the natural logarithm. For non-negative sequences $\{a_n\}$ and $\{b_n\}$, the notation $a_n \leq \cO(b_n)$ (equivalently, $b_n \geq \Omega(a_n)$) indicates the existence of a constant $C > 0$ such that $a_n \leq C b_n$ for all $n > 0$, whereas $a_n = \Theta(b_n)$ signifies the existence of constants $C_1, C_2 > 0$ satisfying $C_1 b_n \leq a_n \leq C_2 b_n$ for all $n > 0$. The term $\Theta_d(1)$ serves as a variant of $\Theta(1)$, emphasizing that both $C_1$ and $C_2$ are independent of $d$. We use $a \propto b$ to denote $a = \Theta(b)$, and $a_n = o(b_n)$ for $\lim_{n \to \infty} a_n / b_n = 0$.

\subsection{AdamW Optimizer}

% \todoq{it is better to add the citation to Adam in proper place. I don't see this citation throughout the paper}

Let $\mW$ denote a parameter tensor, and let $\cL$ be the loss function evaluated on a sampled mini-batch at the current step. 
In this paper, we analyze the model at a fixed point \textbf{in time}, in a static fashion. Therefore, we do not include the timestep throughout the analysis.

We denote by $\mG \triangleq \nabla_{\mW} \cL$ the gradient of the loss with respect to the parameter tensor $\mW$. AdamW \citep{Adamw}, a variant of Adam \citep{adam} with decoupled weight decay, maintains the bias-corrected, accumulated first- and second-order moments, $\mM$ and $\mV$, of the gradient $\mG$ using exponential moving-average coefficients $\beta_1$ and $\beta_2$, and updates
\[
  \mW \leftarrow \mW - \eta \big( \hat \mG + \lambda \mW \big),
  \qquad
  \hat \mG \triangleq \mM \oslash \big(\sqrt{\mV} + \eps\big).
\]
% We refer to $\hat \mG$ as the \textbf{preconditioned gradient} of $\cL(\cdot)$ at $\mW$.

Throughout this paper, we neglect the stabilizer $\eps > 0$, as it is typically set to an insignificantly small value.

\subsection{Parameter Classes}
Denote $d$ as the \textit{model width} (embedding dimension). We study the scaling rule with respect to $d$ throughout this work while keeping other architectural parameters fixed (e.g., model depth, number of MHA (multi-head attention) heads, and FFN (feed-forward network) expansion coefficient). Based on how the parameter shapes scale with $d$, we divide the parameters into two groups:
\begin{itemize}
\item \textbf{Vector-like.} This group contains parameters whose number of entries scales linearly with the model width. Examples include the embedding layer (shape $E \times d$ with fixed vocabulary size $E$), RMSNorm gains (shape $d$), and other one-dimensional parameters. We denote a representative vector parameter by $\mW \in \RR^{1\times d}$.

\item \textbf{Matrix-like.} This group contains parameters whose number of entries scales quadratically with the model width. These include all dense projections in MHA and FFN blocks. In LLaMA-style models, this includes the attention projections $\mW_Q,\mW_K,\mW_V,\mW_O\in\RR^{d\times d}$ in MHA, and the linear projections $\mW_{\mathrm{gating}}, \mW_{\mathrm{in}}\in\RR^{m\times d}$ and $\mW_{\mathrm{out}}\in\RR^{d\times m}$ in FFN, with a fixed expansion ratio $m/d$. When a matrix is rectangular with dimensions $(\kappa d) \times d$ for constant $\kappa$, we treat it as $\kappa = \Theta(1)$ square $d\times d$ blocks. We denote a representative square block by $\mW\in\RR^{d\times d}$.
\end{itemize}
The two groups exhibit different behaviors as the model width $d$ scales. Thus, ideal \textbf{transferable parameterization schemes} should assign distinct hyperparameter scalings to each group.

\subsection{Maximal-update Parameterization}
Maximal-update Parameterization ($\mu$P) principally suggests that \textit{per-step functional changes} should be width-invariant. Specifically, \citet{yang2022tensor} suggest scaling hyperparameters such that the features and their updates after a single step of gradient descent remain invariant to the model width $d$:
\begin{condition}[Desideratum of $\mu$P]
\label{cond:mup-desideratum}
    Let $\vy = f(\vx; \mW)$ be a sublayer with input $\vx$, output $\vy$, and weight matrix $\mW$. Denote by $\Delta \vy$ the change in the output after one gradient update step. The goal is to ensure that
    \[
        \frac{\|\vy\|_{\rms}}{\|\vx\|_{\rms}} = \Theta_d(1) \qquad \text{and} \qquad \frac{\|\Delta \vy\|_{\rms}}{\|\vx\|_{\rms}} = \Theta_d(1).
    \]
\end{condition}

By analyzing the model dynamics around parameter initialization, where parameter magnitudes are dominated by the initial variance, \citet{yang2022tensor} suggest that one should scale the two types of parameters differently based on how their shapes scale with model width. Their prescription yields:
\begin{table}[!ht]
\centering
\begin{tabular}{lcc}
\toprule
Class & Initial variance & Learning rate \\
\midrule
Vector-like  & $\sigma_1^2 = \Theta_d(1)$   & $\eta_1 = \Theta_d(1)$ \\
Matrix-like  & $\sigma_2^2 = \Theta(d^{-1})$ & $\eta_2 = \Theta(d^{-1})$ \\
\bottomrule
\end{tabular}
\caption{$\mu$P scaling rules by parameter class.}
\label{tab:mup-scaling}
\end{table}

% \todoq{is this categorization fully correct? what about output layer(lm head), whose lr is also 1/d?}

We remark that for matrix-like sublayers, the initial variance follows fan-in scaling \citep{glorot2010understanding,he2015delving}, and the learning rate $\eta_2 \propto d^{-1}$ enforces width-invariant per-step functional changes \citep{yang2022tensor}.
The scaling rule further suggests scaling the attention temperature by $1 / d_k$, inversely proportional to the head dimension $d_k$, which itself scales with model width $d$. It also recommends scaling the vocabulary readout as $z_i = \langle \ve_i, \vy \rangle / d$ to keep the logit scale invariant.

In this paper, we complement prior research by studying the scaling rule for the weight decay coefficient $\lambda$ with respect to the model width $d$.

\section{Robust Scaling Rule by Propoer Weight Decay Scaling}

In this section, we present the weight decay scaling rule for robust hyperparameter tuning. The key idea is based on how weight decay interacts with the framework of $\mu$P.

Recent works \citep{kosson2023rotational,defazio2025gradients} show that, in the presence of weight decay and for homogeneous models, the training dynamics of AdamW enter a stable regime governed not by initialization or training timestep, but by the learning rate and weight decay. In this regime, when the hyperparameters of AdamW are held fixed, the weight norm of the weight matrix stabilizes and lies on a sphere, where each training update behaves like a rotation.

Specifically, when training a model parameter $\mW$ in a homogeneous sublayer using AdamW with learning rate $\eta$ and weight decay $\lambda > 0$, the root-mean-square norm of the weight matrix $\mW$ quickly converges to
\[
    \|\mW\|_{\rms} = \Theta_d \bigg( \sqrt{\frac{\eta}{\lambda}} \bigg),
\]
independent of the initialization. This suggests that, instead of tuning the initial variance, we should tune the weight decay $\lambda$ to ensure that the sublayer gain $\|\vy\|_{\rms}/\|\vx\|_{\rms}$ remains invariant across model width.

Consider a linear layer in which $\vy = f(\vx; \mW) \triangleq \mW \vx$. The sublayer gain can be written as
\begin{align} \label{eq:sublayer-gain}
    \frac{\|\vy\|_{\rms}}{\|\vx\|_{\rms}} = \|\mW\|_{\rms} \cdot \rho(d),
\end{align}
where $\rho(d)$ is the alignment factor, which is influenced by both the distribution of the singular value spectra of $\mW$ and the alignment between the spectra and the input vector $\vx$. To make the sublayer gain invariant across different model widths, it is sufficient to determine the power law of $\rho(d)$.

\subsection{Matching Weight Matrix Spectra under Fixed Model Width}

\begin{figure}[!t]
    \centering
    \includegraphics[width=\linewidth]{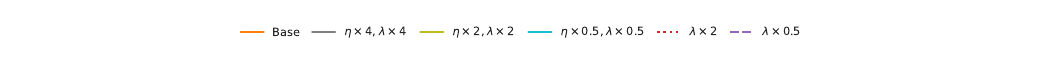}
    \includegraphics[width=0.49\linewidth]{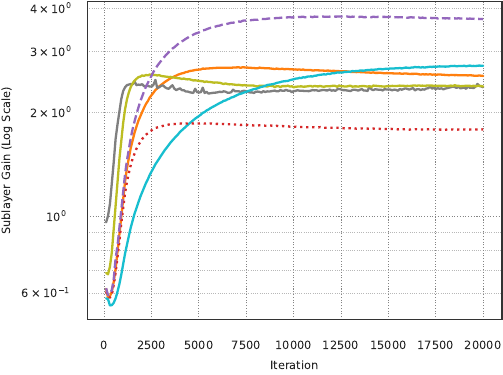}
    \hspace{0.007\linewidth}
    \includegraphics[width=0.49\linewidth]{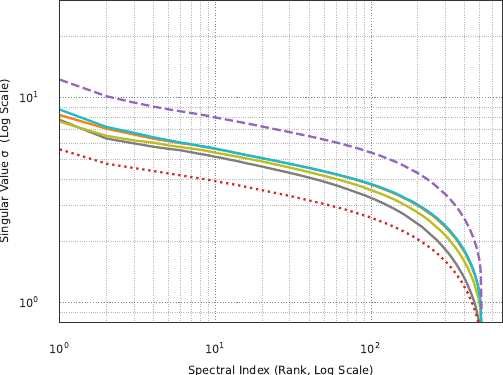}
    \caption{Statistics of the FFN weight matrices under AdamW with various learning rates $\eta$ and weight decay values $\lambda$. The plotted lines are averaged across matrix-like sublayers from all blocks.
    \textbf{Left}: Sublayer gain $\|\vy\|_{\rms}/\|\vx\|_{\rms}$ during training.
    \textbf{Right}: Singular value spectrum $\sigma_i(\mW)$ of the final weight matrices. The singular values are sorted in descending order; the horizontal axis shows the spectral index, and the vertical axis shows the corresponding singular value.}
    \label{fig:spectra-LR-WD-sweep}
\end{figure}
% \todoq{In Figure 1, how can we tell the $\sqrt{\eta/\lambda}$ pattern?}

Although the alignment factor $\rho(d)$ cannot be directly determined beforehand, our key observation is that the ratio between learning rate and weight decay, $\sqrt{\eta/\lambda}$, predominantly controls the overall magnitude of the singular value spectra, whereas their shape remains nearly unchanged. Consequently, sublayer-gain invariance can be achieved by matching the spectra, which in turn can be realized through an appropriate scaling rule for the weight decay.

In our experiment, we train LLaMA models \citep{touvron2023LLaMA} of width $d=512$ on the FineWeb dataset \citep{penedo2024fineweb} using the AdamW optimizer, without learning-rate annealing but with a short warm-up period. Under these conditions, the models reach the steady state characterized by \emph{rotational equilibrium} \citep{kosson2023rotational,defazio2025gradients}. The complete training configuration is summarized in \Cref{tab:exp-setup2}, where a smaller baseline learning rate, $\eta_{\mathrm{base}} = 5.0\times10^{-3}$, is adopted to prevent divergence when scaling $\eta$ upward.

Every 1000 steps, we sample a batch from the training data and run a forward pass. For each linear sublayer $\vy = \mW\vx$, we record the input scale $\|\vx\|_{\rms}$ and output scale $\|\vy\|_{\rms}$ of the data batch, and report their ratio $\|\vy\|_{\rms} / \|\vx\|_{\rms}$ as the \emph{sublayer gain}. In addition, we compute the singular-value spectra of all matrix-like parameters at the end of training.

The results, shown in \Cref{fig:spectra-LR-WD-sweep}, indicate that proportional scaling of the learning rate $\eta$ and weight decay $\lambda$ leaves both the sublayer gains and singular values nearly invariant. In contrast, doubling the weight decay $\lambda$ leads to a uniform down-scaling of both quantities by approximately $\sqrt{2}$. Overall, the magnitude of the spectra grows proportionally to $\sqrt{\eta/\lambda}$, while their shape remains stable across runs.

\begin{fact}
    In the steady state of AdamW training, fixing the model width $d$, the singular values satisfy
    \[
        \sigma_i(\mW) \propto \sqrt{\frac{\eta}{\lambda}},
    \]
    implying that proportional scaling of the learning rate and weight decay preserves the sublayer gain.
\end{fact}

We remark that numerically estimating the scaling rule of spectrum is difficult because no single scalar captures spectrum magnitude well. The operator norm $\|\mW\|_{\mathrm{op}}$ reflects only the largest singular value, ignoring the rest. The Frobenius norm $\|\mW\|_{\mathrm{F}} \triangleq d \cdot \|\mW\|_{\rms}$ aggregates all singular values but collapses both the decay profile and potential sparsity into one number. Moreover, the spectral tail has little impact on signal amplification and is less relevant for satisfying width-invariant gain. In practice, direct visualization of the spectrum remains the most reliable diagnostic.

\subsection{Matching Spectra Induces Sublayer Gain Invariance}
\label{sec:matching-spectra}

\begin{figure}[!t]
    \centering
    \includegraphics[width=\linewidth]{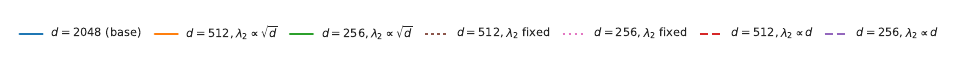}
    \includegraphics[width=0.49\linewidth]{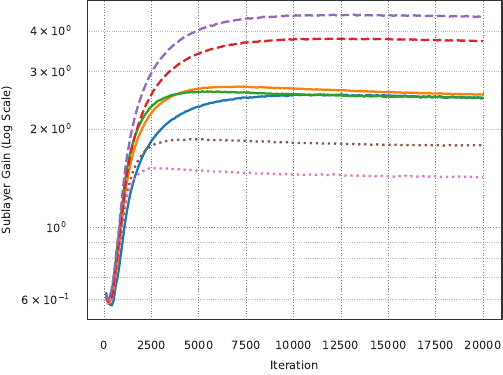}
    \hspace{0.007\linewidth}
    \includegraphics[width=0.49\linewidth]{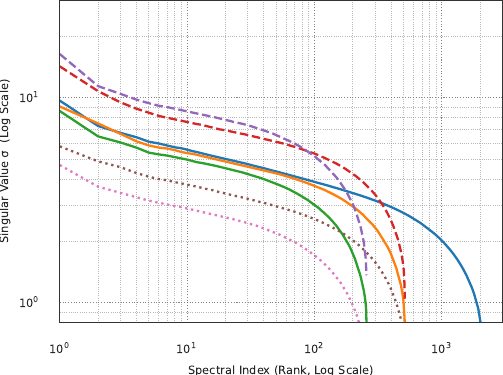}
    \caption{Statistics of the FFN weight matrices under AdamW with various matrix-like weight decay scalings $\lambda_2$, and learning rate scaling specified in \Cref{tab:mup-scaling}, where $\eta_1 = \Theta_d(1)$ is used for vector-like parameters and $\eta_2 \propto 1 / d$ for matrix-like parameters. The plotted lines are averaged across matrix-like sublayers from all blocks.
    \textbf{Left}: Sublayer gain $\|\vy\|_{\rms}/\|\vx\|_{\rms}$ during training. 
    \textbf{Right}: Singular value spectrum $\sigma_i(\mW)$ of the final weight matrices. 
    Alignment is observed when the weight decay scaling follows $\lambda_2 \propto \sqrt{d}$.
    }
    \label{fig:spectra-WD-sweep}
\end{figure}

Next, we sweep the scaling rules of weight decay to examine how the singular value spectra evolve with model width $d$. In the LLaMA training configuration under study, we find that the alignment factor $\rho(d)$ approximately follows $\rho(d) \propto d^{0.75}$, as the singular value spectra align across different widths within matrix-like parameters when $\sqrt{\eta_2 / \lambda_2} \propto d^{-0.75}$. As shown in \Cref{fig:spectra-WD-sweep}, this scaling results in consistent spectral alignment and preserves amplification behavior across model widths. In contrast, alternative weight decay rules lead to noticeable deviations in the spectra. These results suggest that gain invariance is effectively maintained under the proposed parameterization scheme.

To summarize the empirical trend, we state the following observation:
\begin{fact}
    Consider a \emph{matrix-like} linear sublayer $\vy = f(\vx; \mW) \triangleq \mW \vx$ in a Transformer, where $\vx$ denotes the input, $\vy$ the output, and $\mW$ the weight matrix. When the sublayer is trained using AdamW with learning rate $\eta$, weight decay $\lambda$, and all other hyperparameters fixed, scaling the model width $d$ while maintaining
    \[
        \sqrt{\frac{\eta}{\lambda}} \propto d^{-0.75}
    \]
    aligns the magnitude of the weight matrix top singular values $\sigma_i(\mW)$ across widths and keeps the sublayer gain $\|\vy\|_{\rms} / \|\vx\|_{\rms}$ approximately invariant.
\end{fact}

By combining this rule with the learning-rate scaling of $\mu$P, the initialization scaling rule that ensures proper behavior at initialization, and the common practice of disabling weight decay for vector-like parameters, we obtain the following parameterization scheme:
\begin{table}[!ht]
\centering
\begin{tabular}{lccc}
\toprule
Class & Initial variance & Learning rate & Weight decay \\
\midrule
Vector-like  & $\sigma_1^2 = \Theta_d(1)$     & $\eta_1 = \Theta_d(1)$     & $\lambda_1 = 0$ \\
Matrix-like  & $\sigma_2^2 = \Theta(d^{-1})$ & $\eta_2 = \Theta(d^{-1})$  & $\lambda_2 = \Theta(\sqrt{d})$ \\
\bottomrule
\end{tabular}
\caption{Our proposed layerwise scaling rules by parameter classes.}
\label{tab:new-scaling}
\end{table}

\subsection{Tuning Weight Decay Enables Hyperparameter Transfer}
\begin{figure}[!t]
    \centering
    \includegraphics[width=0.475\linewidth]{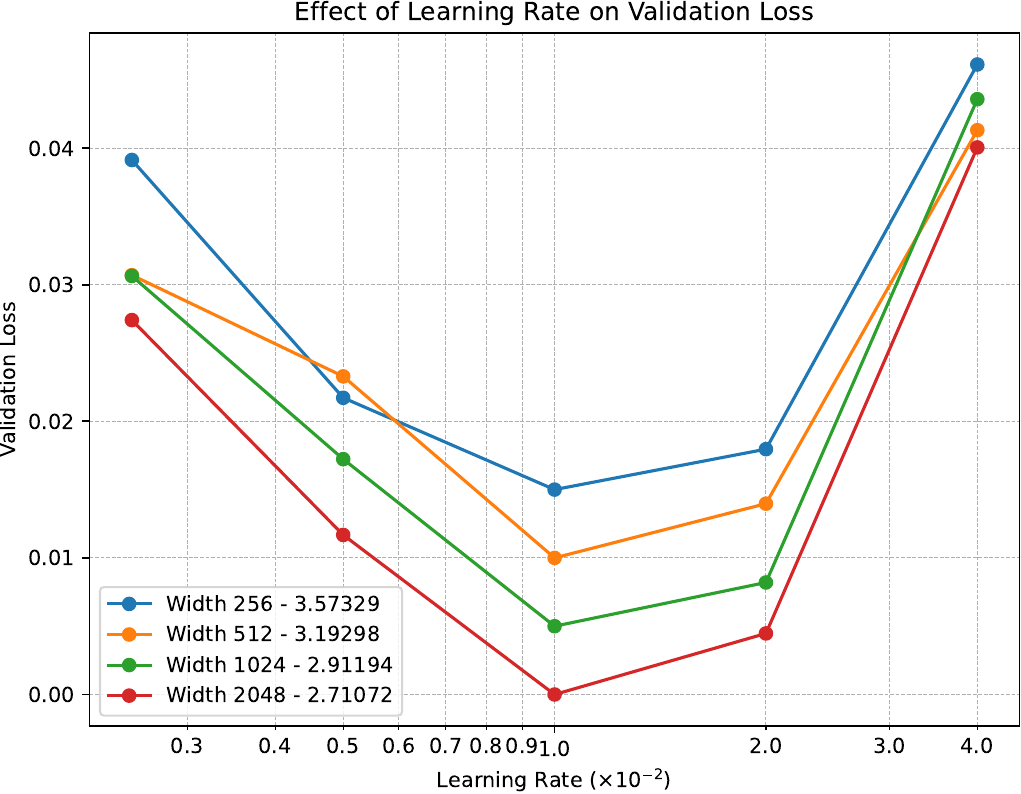}
    \includegraphics[width=0.475\linewidth]{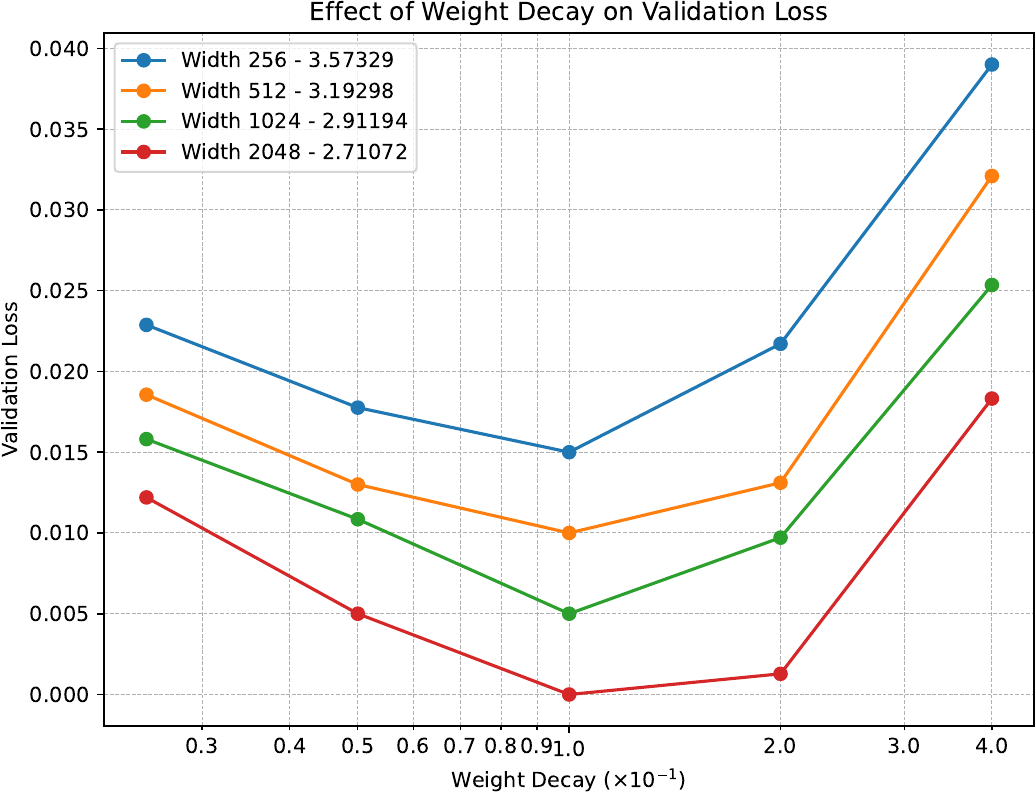}
    \caption{Transfer of the optimal base learning rate $\eta_{\mathrm{base}}$ \textbf{(Left)} and weight decay $\lambda_{\mathrm{base}}$ \textbf{(Right)} across model widths. Each curve shows the loss landscape for a specific width $d$, with minima aligned after scaling according to \Cref{tab:new-scaling}. For visualization, all curves are vertically shifted by constant offsets so that the losses are directly comparable across widths. The alignment of minima indicates that the proposed parameterization enables consistent hyperparameter transfer across scales.}
    \label{fig:HP-transfer}
\end{figure}

We evaluate the new scaling rule by sweeping the base learning rate $\eta_{\mathrm{base}}$ and base weight decay $\lambda_{\mathrm{base}}$ across models of varying widths. We train LLaMA-style models on FineWeb with widths ranging from $d = 256$ (approximately 19M parameters) up to $d = 2048$ (approximately 500M parameters). 
The hyperparameters are scaled such that $\eta_1 = \eta_2 = \eta_{\mathrm{base}}$ and $\lambda_2 = \lambda_{\mathrm{base}}$ at the base model width $d_{\mathrm{base}} \triangleq 256$, and the remaining values are scaled according to \Cref{tab:new-scaling}.
Each model is trained for 20{,}000 steps using cosine learning rate annealing, which decays to $0.01\times$ the peak value after a linear warm-up of 1{,}000 steps. The configuration is summarized in \Cref{tab:exp-setup2}. As shown in \Cref{fig:HP-transfer}, the proposed parameterization scheme enables consistent transfer of the optimal base learning rate $\eta_{\mathrm{base}}$ and weight decay $\lambda_{\mathrm{base}}$ across model widths.

This consistency supports two practical scaling strategies:
\begin{itemize}
    \item \textbf{Base-to-Target Transfer.} Following the $\mu$P parameterization view, select a small base width $d_{\mathrm{base}}$, perform a local hyperparameter sweep, and compute the corresponding hyperparameters for a larger target width $d_{\mathrm{target}}$ using the derived scaling rules in \Cref{tab:new-scaling}. This approach will lead to learning rates of matrix-like parameters that are comparably smaller than those of vector-like ones.
    
    \item \textbf{Proxy-to-Target Scaling.} Set the base width $d_{\mathrm{base}} \triangleq d_{\mathrm{target}}$ to the target model width, and perform a hyperparameter search at a small width $d_{\mathrm{proxy}}$ with layerwise-tuned learning rates according to the new scaling rule in \Cref{tab:new-scaling}. In this way, we can use standard parameterization for the target model while benefiting from zero-shot hyperparameters obtained via proper hyperparameter scaling.
\end{itemize}

We remark that changing the base width $d_{\mathrm{base}}$ can be viewed as redefining the reference point for layerwise scaling, which in turn adjusts the learning-rate ratio $\eta_1 / \eta_2$ between vector-like and matrix-like parameters.

\subsection{Tradeoff between Learning Rate and Weight Decay}

\begin{figure}[!t]
    \centering
    \includegraphics[width=0.6\linewidth]{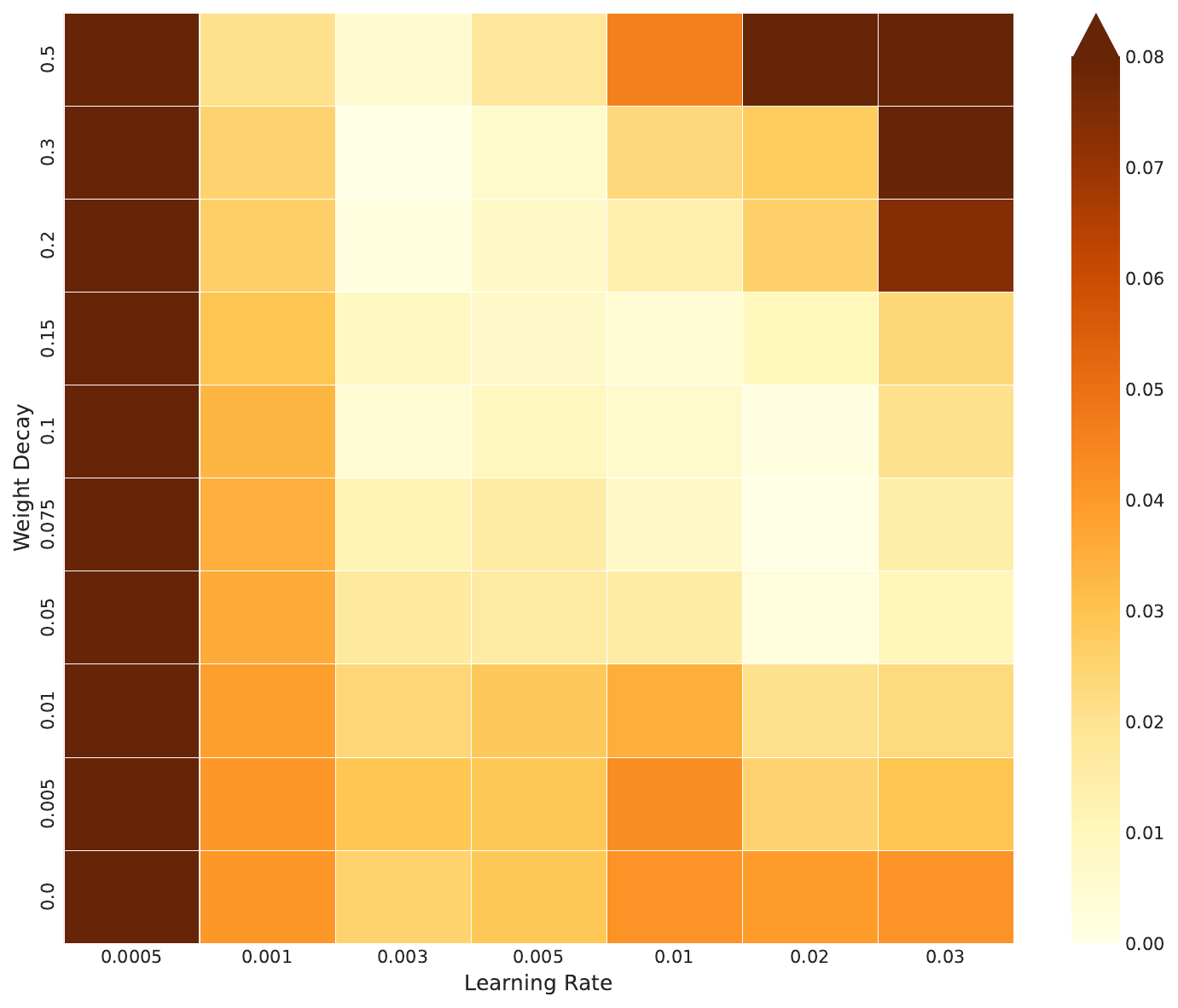}
    \caption{Validation loss differences across learning-rate and weight-decay pairs. Loss values are measured relative to the optimal configuration  $(\eta_{\mathrm{base}} = 0.02,\; \lambda_{\mathrm{base}} = 0.075)$. A clear diagonal ridge of near-optimal points reveals that increasing $\lambda$ requires reducing $\eta$, confirming their strong correlation. 
Both axes are spaced approximately logarithmically.}
    \label{fig:2d-search}
\end{figure}

\Cref{fig:2d-search} shows that the optimal learning rate decreases as the weight decay increases, forming an approximately diagonal ridge of near-optimal configurations. This ridge indicates that the learning rate and weight decay are not independent hyperparameters but are closely correlated. There is no single learning rate that works uniformly across all weight decays, and vice versa. Similar correlations have been reported by \citet{d2024we}, who also observe that tuning one parameter requires adjusting the other.

More importantly, we find that the models located along this ridge achieve almost identical final losses despite having different weight decays. This suggests that the precise choice of weight decay is less critical once the corresponding optimal learning rate is used. Consequently, a full two-dimensional hyperparameter search could often be unnecessary. In practice, one can heuristically select a weight decay, perform a one-dimensional sweep over the learning rate, and then transfer the resulting configuration to larger models using the $\mu$P scaling rules summarized in \Cref{tab:new-scaling}. All models in this experiment are trained on 5B tokens with 500 warmup steps.

\section{Illustrative Example of Sqrt Weight Decay Scaling}
\label{sec:synthetic-run}

\begin{figure}[!t]
    \centering
    \includegraphics[width=\linewidth]{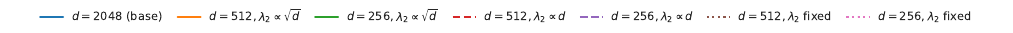}
    \includegraphics[width=0.49\linewidth]{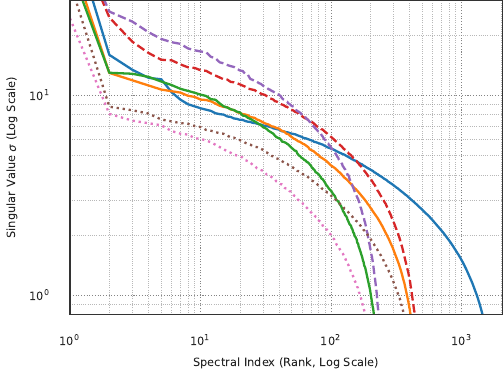}
    \hspace{0.007\linewidth}
    \includegraphics[width=0.49\linewidth]{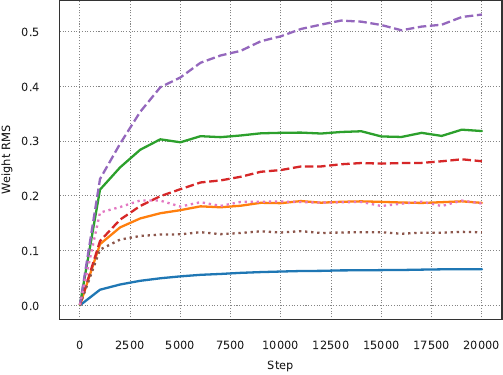}
    % \TODO
    \caption{Statistics of the weight matrix $\mW_{\mathrm{in}}$ under AdamW in the synthetic run with learning rate scaling $\eta \propto 1 / d$ and various scalings of weight decay $\lambda$.
    \textbf{Left}: Singular value spectrum $\sigma_i(\mW)$ of the final weight matrices. When the weight decay follows $\sqrt{\eta / \lambda} \propto d^{-0.75}$, the top singular values are approximately aligned, matching our realistic training results in \Cref{sec:matching-spectra}. 
    \textbf{Right}: During training, the root-mean-square of the weight matrix $\|\mW\|_{\rms}$ converges to a stable value, approximately proportional to $\sqrt{\eta / \lambda}$.}
    \label{fig:synthetic-run}
\end{figure}

In this section, we present an illustrative model using synthetic data, where the square-root weight decay scaling rule can be observed. Although, to our knowledge, this rule has not been previously reported in the literature, we note that it arises naturally from the model architecture rather than the data distribution.

Consider a two-layer FFN $\vy = f(\vx; \mW) \triangleq \mW_{\mathrm{out}} \cdot \varphi(\mW_{\mathrm{in}} \vx)$ with weight matrices $\mW_{\mathrm{in}}, \mW_{\mathrm{out}} \in \RR^{d \times d}$ and ReLU activation function $\varphi$. We train the model on synthetic data, where both the input vector $\vx \sim \cN(0, \mI_d)$ and the upstream gradient $\nabla_{\vy} \cL \sim \cN(0, \mI_d)$ are independently drawn from standard normal distributions. In this setup, by the chain rule, the gradients of the sublayers are given by
\[
    \nabla_{\mW_{\mathrm{out}}} \cL = \nabla_{\vy} \cL \cdot \varphi(\mW_{\mathrm{in}} \vx)^\top, 
    \qquad 
    \nabla_{\mW_{\mathrm{in}}}  \cL = \big((\mW_{\mathrm{out}}^\top \nabla_{\vy} \cL) \odot \varphi'(\mW_{\mathrm{in}} \vx)\big) \vx^\top.
\]

We train the model across varying widths $d$, from $256$ to $2048$, using AdamW with a learning rate scaled as $\eta \propto 1 / d$, following the $\mu$P scaling rule for matrix-like sublayers and the proposed weight decay scaling rule $\lambda \propto \sqrt{d}$. The model is trained for $20{,}000$ steps without learning rate annealing, reaching the steady state characterized by rotational equilibrium \citep{kosson2023rotational,defazio2025gradients}. The AdamW hyperparameters match those used in our main LLaMA experiment in \Cref{fig:spectra-WD-sweep}, specifically for matrix-like layers.

As shown in \Cref{fig:synthetic-run,fig:synthetic-run-other}, the top singular value spectra of the weight matrices align under the weight decay scaling rule $\lambda \propto \sqrt{d}$, consistent with the experimental results on LLaMA in \Cref{fig:spectra-WD-sweep}. This further suggests that the observed weight decay scaling rule arises naturally in training regimes with very high variance, such as Transformers trained for next-token prediction.

\section{Conclusion}

In this paper, we study the weight decay scaling rule of $\mu$P in light of the steady-state dynamics characterized by rotational equilibrium \citep{kosson2023rotational}. This result complements the original $\mu$P framework, which guarantees early-stage sublayer gain by scaling the initialization variance to match the sublayer gain during the steady stage of training. 

We find that for matrix-like parameters in linear layers, setting the ratio between learning rate and weight decay as $\sqrt{\eta/\lambda} \propto d^{-0.75}$, where $d$ denotes model width, enables the top singular value to align across model widths. This, in turn, implies that sublayer gain remains approximately invariant across different widths. 
The observation leads to a layerwise scaling rule: assign vector-like parameters a learning rate of $\eta_1 = \Theta_d(1)$ and a weight decay of $\lambda_1 = 0$, independent of model width. For matrix-like parameters, use a learning rate scaling as $\eta_2 \propto d^{-1}$ and weight decay scaling as $\lambda_2 \propto \sqrt{d}$. This layerwise scaling rule enables zero-shot hyperparameter transfer, in contrast to traditional scaling rules that require an optimal learning rate sweep at each model width.
An illustrative example is also provided to demonstrate that square-root weight decay scaling can be observed in minimal models. The success of this approach may hint at mean-field-like behavior in large models, where training dynamics are well captured by random matrix theory.

We remark that the concurrent work \citep{filatov2025optimal} empirically observes that (near-)optimal training hyperparameters equalize operator norms across widths, mirroring our finding that maintaining width-invariant sublayer gain yields robust transfer. Their mechanism differs from ours: rather than tuning weight decay, they adjust the batch size to match sublayer gain. Since batch size primarily controls the gradient-noise scale and thus acts as an implicit regularizer \citep{welling2011bayesian,mandt2017stochastic}, while weight decay constitutes explicit regularization , the two observations are closely related. Our explicit-regularization view leads to cleaner, layerwise scaling prescriptions (e.g., vector- vs.\ matrix-like parameters) and yields a more directly interpretable rule for extrapolating across widths than batch-size tuning alone.

\textbf{Scope.}  
Our results apply to AdamW and LLaMA architectures with a fixed number of heads and FFN ratio. It is not obvious whether the observed scaling rule $\lambda_2$ is universal across \emph{all} architectures: mixture-of-experts architectures, alternatives to self-attention, or other architectural choices might alter the scaling factor, and this would be interesting to study. However, we believe that inspecting sublayer gain using singular value spectra and attempting to match the top singular value spectra is a transferable procedure that may be adopted for extended research.

\textbf{Outlook.}  
Future work includes extending the research to other optimizers (e.g., SGD with momentum, Adafactor), mixture-of-experts and structured-sparse models, and regimes where batch size or training tokens grow with width. It would also be a promising direction to study how to scale hyperparameters when increasing model depth. Developing a predictive link between data distribution, optimizer statistics, and spectral shape could turn the empirical law for $\sqrt{\eta/\lambda}$ into a principled theory for steady-state transfer. Our illustrative example of a two-layer feed-forward network can be seen as a step in this direction.

\textbf{Perspective.}  
We advocate studying LLM training as a dynamical physical system, using tools from dynamical systems and statistical physics. In this view, theory plays a role analogous to fluid mechanics: not exact in every detail, but predictive at the right scales. In practice, this means privileging models that explain and forecast observed phenomena, e.g., training at the edge of stability \citep{cohen2021gradient,arora2022understanding}, rotational steady states under decoupled weight decay \citep{Adamw,kosson2023rotational}, and noise-driven effects captured by stochastic-thermodynamic views of SGD \citep{welling2011bayesian,mandt2017stochastic}, over exact-but-fragile formalisms. Echoing Box's dictum that ``all models are wrong, but some are useful'' \citep{box1976science}, our aim is to develop useful, testable models of steady-state training dynamics that complement mathematical analysis. This work is a step in that direction.

\bibliographystyle{plainnat}
\bibliography{refs}

\clearpage
\appendix

{\huge Appendix}

\section{Hyperparameter Table}
\begin{table}[!ht]
\centering
\small
\setlength{\tabcolsep}{6pt}
\renewcommand{\arraystretch}{1.15}
\begin{tabular}{@{}lccc@{}}
\toprule
\textbf{Setting} & \multicolumn{3}{c}{\textbf{Value / Configuration}} \\
\midrule
\multicolumn{4}{@{}l}{{Model \& Data}} \\
\midrule
Architecture & \multicolumn{3}{l}{LLaMA} \\
Dataset & \multicolumn{3}{l}{FineWeb} \\
Compute & \multicolumn{3}{l}{$\mathrm{H200}$ GPU} \\
Epochs & \multicolumn{3}{l}{1} \\
\midrule
\multicolumn{4}{@{}l}{{Architecture}} \\
\midrule
Model width & \multicolumn{3}{l}{Various $d$} \\
Model depth & \multicolumn{3}{l}{8} \\
MHA heads Count & \multicolumn{3}{l}{$16$} \\
FFN expansion ratio & \multicolumn{3}{l}{$8/3$} \\
Context length & \multicolumn{3}{l}{1024} \\
Attention scaling & \multicolumn{3}{l}{$d_{\mathrm{base}}^{-1/2} \cdot m_d^{-1}$} \\
\midrule
\multicolumn{4}{@{}l}{{AdamW hyperparameters}} \\
\midrule
Beta values & \multicolumn{3}{l}{$(\beta_1, \beta_2) = (0.9, 0.95)$} \\
Epsilon & \multicolumn{3}{l}{$\eps=10^{-8}$} \\
\multirow{2}{*}{Schedule} & \multicolumn{3}{l}{1000 steps linear learning rate warmup} \\
& \multicolumn{3}{l}{cosine anneal to $0.01\times$ the peak learning rate, if annealed} \\
\midrule
\multicolumn{4}{@{}l}{{Training setup}} \\
\midrule
Grad-norm clip & \multicolumn{3}{l}{1.0} \\
Dropout & \multicolumn{3}{l}{0} \\
Batch size & \multicolumn{3}{l}{480} \\
Steps & \multicolumn{3}{l}{20,000} \\
Precision & \multicolumn{3}{l}{BF16} \\
\midrule
\multicolumn{4}{@{}l}{{Parameterization}} \\
\midrule
\textbf{Setting} & \textbf{Base} & \textbf{Type I (vector)} & \textbf{Type II (matrix)} \\
\midrule
Initialization std 
& $\sigma_{\mathrm{base}} = 2.0 \times 10^{-2}$
& $\sigma_1=\sigma_{\mathrm{base}}$
& $\sigma_2=\sigma_{\mathrm{base}}\cdot m_d^{-1/2}$ \\
Learning rate
& Default $\eta_{\mathrm{base}} = 1.0 \times 10^{-3}$
& $\eta_1=\eta_{\mathrm{base}}$
& $\eta_2=\eta_{\mathrm{base}}\cdot m_d^{-1}$ \\
Weight decay
& Default $\lambda_{\mathrm{base}} = 1.0 \times 10^{-1}$
& $\lambda_1=0$
& $\lambda_2=\lambda_{\mathrm{base}}\cdot m_d^{1/2}$ \\
\bottomrule
\end{tabular}
\caption{Default experimental setup of experiments. Here $m_d\triangleq d/d_{\mathrm{base}}$ with $d_{\mathrm{base}}=256$.}
\label{tab:exp-setup2}
\end{table}

\clearpage
\section{Other Figures}

\begin{figure}[!ht]
    \centering
    \includegraphics[width=\linewidth]{figures/figure_5/proj_svd_legend.pdf}
    \includegraphics[width=0.49\linewidth]{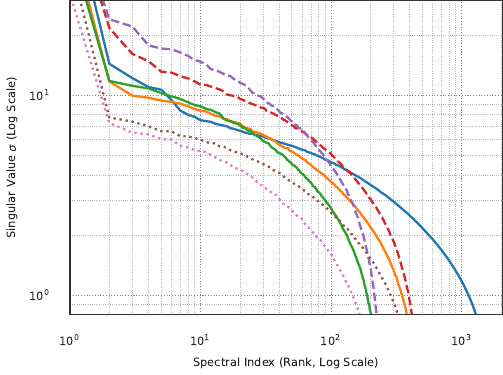}
    \hspace{0.007\linewidth}
    \includegraphics[width=0.49\linewidth]{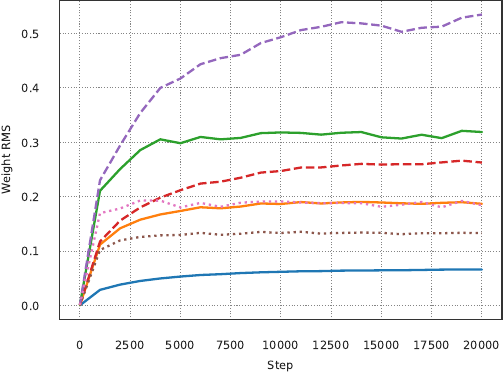}
    % \TODO
    \caption{Statistics of the weight matrix $\mW_{\mathrm{out}}$ in the synthetic run, complementary to \Cref{fig:synthetic-run}.
    }
    \label{fig:synthetic-run-other}
\end{figure}

\end{document}